\documentclass[10pt, a4paper]{article}
\usepackage{lrec}
\usepackage{multibib}
\newcites{languageresource}{Language Resources}
\usepackage{graphicx}
\usepackage{tabularx}
\usepackage{soul}

\usepackage[croatian,english]{babel}

\usepackage{epstopdf}
\usepackage[utf8]{inputenc}

\usepackage{hyperref}
\usepackage{xstring}

\usepackage{color}
\usepackage{colortbl}
\definecolor{mylightgray}{RGB}{219,219,219}

\def\paperDraft{}
\ifdefined\paperDraft
 \def\mpcomment#1{{\color{green}[Matt: \textit{#1}]}}
 \def\csacomment#1{{\color{blue}[Carlos: \textit{#1}]}}

\else
 \def\mpcomment#1{}
 
 \def\csacomment#1{}
 
\fi

\title{CoSimLex: A Resource for Evaluating Graded Word Similarity in Context}

\name{Carlos S. Armendariz$^{\ast}$, Matthew Purver$^{\ast\dagger}$,\\
{\bf \large Matej Ul\v{c}ar$^{\ddagger}$, Senja Pollak$^{\dagger}$, Nikola Ljubešić$^{\dagger}$, \large Marko Robnik-\v{S}ikonja$^{\ddagger}$,}\\ 
{\bf \large Mark Granroth-Wilding$^{\diamondsuit}$, Kristiina Vaik$^{\mathsection}$}}

\address{
$^{\ast}$Cognitive Science Research Group, Queen Mary University of London, London, UK \\
        \{c.santosarmendariz, m.purver\}@qmul.ac.uk\\
$^{\dagger}$Department of Knowledge Technologies, Jožef Stefan Institute, Ljubljana, Slovenia\\
        \{senja.pollak, 	nikola.ljubesic\}@ijs.si\\
$^{\ddagger}$University of Ljubljana, Faculty of Computer and Information Science, Slovenia\\
    \{matej.ulcar, marko.robnik\}@fri.uni-lj.si\\
$^{\diamondsuit}$Department of Computer Science, University of Helsinki, Finland\\
    mark.granroth-wilding@helsinki.fi\\
$^{\mathsection}$Department of Data Analysis, Texta, Estonia\\
    kristiina.vaik@ut.ee\\}

\abstract{
State of the art natural language processing tools are built on context-dependent word embeddings, but no direct method for evaluating these representations currently exists. Standard tasks and datasets for intrinsic evaluation of embeddings are based on judgements of similarity, but ignore context; standard tasks for word sense disambiguation take account of context but do not provide continuous measures of meaning similarity. This paper describes an effort to build a new dataset, CoSimLex, intended to fill this gap. Building on the standard pairwise similarity task of SimLex-999, it provides context-dependent similarity measures; covers not only discrete differences in word sense but more subtle, graded changes in meaning; and covers not only a well-resourced language (English) but a number of less-resourced languages. We define the task and evaluation metrics, outline the dataset collection methodology, and describe the status of the dataset so far. \\ \newline 
 }

\begin{document}

\maketitleabstract

\section{Introduction}

Recent work in language modelling and word embeddings has led to a sharp increase in use of context-dependent models such as ELMo \cite{PetersEtAl18ELMO} and BERT \cite{devlin2019bert}. These models, by providing representations of words which depend on the surrounding context, allow us to take account of the effects not only of discrete differences in word sense but of the more graded effects of context. However, evaluation of these models has generally been in terms of either their performance as language models, or their effect on downstream tasks such as sentiment classification \cite{PetersEtAl18ELMO}: there are few resources available which allow  evaluation in terms of the properties of the embeddings themselves, or in terms of their ability to model human perceptions of meaning. There are established methods to evaluate word embedding models intrinsically via their ability to reflect human similarity judgements (see e.g.\ WordSim-353 \cite{finkelstein2002placing} and SimLex-999 \cite{HillEtAl15SimLex}) or model analogies \cite{mikolov2013efficient}; however, these have generally ignored context and treated words in isolation. The few that do provide context (e.g.\ SCWS \cite{huang2012improving} and WiC \cite{pilehvar2019wic}) focus on word sense and discrete effects, thus missing some of the effects that context has on words in general, and some of the benefits of context-dependent models. To evaluate current models, we need a way to evaluate their ability to reflect similarity judgements \emph{in context}: how well do they model the effects that context has on word meaning?

In this paper we present our ongoing efforts to define and build a new dataset that tries to fill that gap: \textbf{CoSimLex} \citelanguageresource{cosimlex}. CoSimLex builds on the familiar pairwise, graded similarity task of SimLex-999, but extends it to pairs of words as they occur in context, and specifically provides two different shared contexts for each pair of words. This will provide a dataset suitable for intrinsic evaluation of state-of-the-art contextual word embedding models, by testing their ability to reflect human judgements of word meaning similarity in context, and crucially, the way in which this varies as context is changed. It goes beyond other existing context-based datasets by taking the \emph{gradedness} of human judgements into account, thus applying not only to polysemous words, or words with distinct senses, but to the phenomenon of context-dependency of word meaning in general. The dataset is also multi-lingual, and includes three less-resourced European languages: Croatian, Finnish and Slovene.
It is to be used as the gold standard for evaluation of a task at SemEval2020: Task 3, Graded Word Similarity in Context.\footnote{https://competitions.codalab.org/competitions/20905}

\section{Background}\label{sec:bg}

From the outset, our main motivation for the development of this dataset came from an interest in the cognitive and psychological mechanisms by which context affects our perception of the meaning of words. There have been many different ways in the literature to look at this phenomenon, which lie in the intersection of several different fields of research, and a detailed discussion of the different approaches to this problem is out of the scope of this paper; here, we present two of the most prominent ideas that helped define what we were trying to capture, and made an impact in the design of the dataset and its annotation process. We then look at previous datasets that deal with similarity in context.

\subsection{Contextual Modulation}
Within the field of lexical semantics, \newcite{cruse1986lexical} proposed an interesting compromise between those linguists that saw words as associated with a number of discrete senses and those that thought that the perceived discreteness of lexical senses is just an illusion. He distinguishes two different manners in which sentential context modifies the meaning of a word. First, the context can select for different discrete senses; if that is the case, the word is described as \emph{ambiguous}, and the process is referred as \textbf{contextual selection of senses}. This effect is well known, and is the basis of many word-sense disambiguation tasks.  

\newcounter{list_counter}
\begin{enumerate}
    \item\label{ex:bank1} We finally reached the bank.
    \item\label{ex:bank2} At this point, the bank was covered with brambles.
    \setcounter{list_counter}{\value{enumi}}
\end{enumerate}

In example (\ref{ex:bank1}), the  word \emph{bank} can have the \emph{financial} or \emph{riverbank} sense; and here, the context doesn't really help us select the correct sense. This creates some tension on the part of the reader: 
we need to select a sense in order for the sentence to properly work, and without this we may feel that the sentence has not been fully understood. This is an example of \emph{ambiguity}. In example (\ref{ex:bank2}), in contrast, the context makes one of the senses more \emph{normal} than the other. \newcite{cruse1986lexical} sees the evaluation of \emph{contextual normality} as the main mechanism for sense selection. 

The second way in which context can modify the meaning of a word works within the scope of a single sense, modifying it in an unlimited number of ways by \emph{highlighting} certain semantic traits and \emph{backgrounding} others. This process is called \textbf{contextual modulation of meaning}, and the word is said to be \emph{general} with respect to the traits that are being modulated. This effect is by nature not discrete but continuous and fluid, and since every word is \emph{general} to some extent: it can be argued that a word has a different meaning in every context in which it appears.

\begin{enumerate}
	\setcounter{enumi}{\value{list_counter}}
	\item\label{ex:cousin1} Sue is visiting her pregnant cousin.
	\item\label{ex:cousin2} Peter doesn't like his cousin.
	\item\label{ex:butter} Arthur poured the butter into a dish.
	\setcounter{list_counter}{\value{enumi}}
\end{enumerate}

In example (\ref{ex:cousin1}), the context tells us that the cousin is female. The meaning of \emph{cousin} is being \emph{modulated} by the context to promote the ``female'' trait. \emph{Cousin} is a \emph{general} word that includes male and female, but also tall, short, happy and sad cousins. However, as we can see in example (\ref{ex:cousin2}), the absence of information about these traits doesn't produce the type of tension we saw in (\ref{ex:bank1}) above; there is a distinction between  meaning modulation and sense selection.

The last example (\ref{ex:butter}) is another case of \emph{contextual modulation} in which \emph{poured} highlights the ``liquid'' trait for \emph{butter}. It is interesting to notice that in this case not only ``liquid'' is highlighted, related traits like ``warm'' can be highlighted as a consequence.

It seems clear that the contextual selection of senses would modify human judgements of similarity. For example, the word \emph{bank}, when used in a context which selects its financial institution sense, should be scored as more similar to other kinds of financial institution (e.g.\ \emph{building society}) than when in a context which selects the geographic sense of the word. However, we should also expect that a word like \emph{butter}, when contextually modulated to highlight its ``liquid'', ``hot'' and ``frying'' traits, should score more similar to \emph{vegetable oil} than when contextually modulated to highlight its ``animal sourced'', ``dairy'', and ``creamy'' traits. This kind of hypothesis would be testable given a new context-dependent similarity dataset.

Both \emph{sense selection} and \emph{meaning modulation} happen very commonly together, with the same context forcing a sense and then modulating its expression. Many different explanations have been proposed for the emergence of these discrete senses, and some may have their origins in very commonly modulated meaning but, according to Cruse, once a discrete sense is established it becomes something different and follows different rules:

\begin{enumerate}
    \setcounter{enumi}{\value{list_counter}}
    \item\label{ex:gender1} John prefers bitches to dogs.
    \item\label{ex:gender2} John prefers bitches to canines.
    \item\label{ex:gender3} Mary likes mares better than horses.
    \setcounter{list_counter}{\value{enumi}}
\end{enumerate}

Here example (\ref{ex:gender1}) works because one of the discrete senses associated to the word \emph{dog} refers only to male dogs. This cannot be explained by \emph{contextual modulation}: if that was the case, example (\ref{ex:gender2}), which replaces \emph{dog} with \emph{canine}, should also work, as \emph{canine} could be modulated in the same way that \emph{dog} was; and similarly example (\ref{ex:gender3}). However, both seem unnatural at best. The fact that neither \emph{canine} nor \emph{horse} can be modulated in this same way indicates that meaning modulation and sense selection are two, strongly interconnected, but distinctive mechanisms of contextual variability.

A final interesting point about Cruse's view is that he doesn't find the contrast between polysemy and homonymy particularly helpful, and dislikes the use of these terms because they promote the idea that the primary semantic unit is some common lexeme and each of the different senses are just variants of it. He instead believes the primary semantic unit should be the \emph{lexical units}, a union of a single sense and a lexical form, and finds it more useful to look at the contrast between discrete and continuous semantic variability.

\subsection{Salience Manipulation} 
Until now we have looked at contextual variability as an exclusively linguistic phenomenon, a point of view rooted in lexical semantics. We looked at how the context of the sentence affects the meaning of the word. In contrast, cognitive linguistics, and the more specific cognitive semantics, look at language and meaning as a more general expression of human cognition \cite{evans2018}. 

This approach champions concepts, more specifically \emph{conceptual structures}, as the true recipient of meaning, replacing words or lexical units. These linguistic units no longer refer to objects in an external world but to concepts in the mind of the speaker. Words get their meaning only by association with \emph{conceptual structures} in our minds. The process by which we construct meaning is called conceptualisation, an embodied phenomenon based in social interaction and sensory experience. 

Cognitive linguists gravitate to themes that focus on the flexibility and the ability of the interaction between language and conceptual structures to model continuous phenomena, like prototyping effects, categories, metaphor theory and new ways to look at polysemy.
Within the cognitive tradition, the idea of \emph{conceptual spaces}, characterised by \emph{conceptual dimensions},  has been especially influential \cite{gardenfors2000,gardenfors2014geometry}. These dimensions can range from concrete ones like weight, temperature and brightness, to very abstract ones like awkwardness or goodness. Once a domain, or selection of dimensions is established, a concept is defined as a region (usually a convex one) of the conceptual space. An example would be to define the colour \emph{brown} as a region of a space made of the dimensions \emph{Red}, \emph{Green} and \emph{Blue}. This geometric approach lends itself perfectly to model phenomena like prototyping (central point of the region), similarity (distance), metaphor (projection between different dimensions) and, more importantly for our concerns here, fluid changes in meaning due to the effects of context.

\newcite{warglien2015meaning} use conceptual spaces to look at \emph{meaning negotiation} in conversation. They investigate the mechanisms, consciously or unconsciously, employed by the people involved in conversation to negotiate meaning of vague predicates, in order to satisfy the coordination needed for communication. These tools help them to decide areas in which they don't agree as well. All these processes work by manipulating the conceptual dimensions in which meaning is represented. We will refer to them as \textbf{salience manipulation} because their main role is to dynamically rise or lower the perceived importance of certain conceptual dimensions. 

The main mechanism by which speakers can modify salience of conceptual dimensions are the automatic \emph{priming} effects described by, for example, \newcite{pickering2004toward}: mentioning specific words early in the conversation can make the dimensions associated with such words more relevant. 
Speakers can also explicitly try to remove dimensions from the domain in order to promote agreement, or bring in new dimensions by using \emph{metaphoric projections}. Because metaphors can be understood as mappings that transfer structure from one domain to another, they can introduce new dimensions and meaning to the conversation. 

\begin{quote}The lion Ulysses emphasizes Ulysses’ courage but hides his condition of a castaway in Ogiya. Thus metaphors act by orienting communication and selecting dimensions that may be more or less favorable to the speaker. By suggesting that a storm hit the financial markets, a bank manager can move the conversation away from dimensions pertaining to his own responsibilities and instead focus on dimensions over which he has no control. \cite{warglien2015meaning}\end{quote}

\begin{figure*}[h!]
	\begin{tabular}{|llr|} \hline 
		\bf Word1: bank & 
		\bf Word2: money & 
		\bf  \\ \hline
		\bf Context1 & 
		& 
		\bf  \\ 
		\multicolumn{3}{|p{\linewidth}|}{
			Located downtown along the east \textbf{bank} of the Des Moines River ... 
		}\\ \hline
		\bf Context2 & 
		& 
		\bf \\
		\multicolumn{3}{|p{\linewidth}|}{
			This is the basis of all \textbf{money} laundering, a track record of depositing clean money before slipping through dirty money ..
		}\\ \hline
	\end{tabular}
	\caption{Example from the SCWS dataset, the focus is in the different senses of the word \textbf{bank} and there is one independent context per word.}\label{fig:scws}
\end{figure*}

From this perspective, then, the change in meaning is no longer a change in the meaning of a specific word, but a change in the mind of the hearer (or reader), a change in their \emph{mental state} triggered by their interaction with the context. We saw an example of the meaning of the word ``butter'' being \emph{contextually modulated} before, lets see some examples of \emph{salience manipulation} having an effect on the same word:

\begin{enumerate}
	\setcounter{enumi}{\value{list_counter}}
	\item\label{ex:muffins} My muffins were a failure, I should have used butter or margarine instead of olive oil. 
	\item\label{ex:vegan1} Vegan chefs replace animal fats, like butter, with plant based ones like olive oil or margarine.
	\item\label{ex:vegan2} Vegan influencers believe the consumption of animal products is cruel and unnecessary.
	\setcounter{list_counter}{\value{enumi}}
\end{enumerate}

In example (\ref{ex:muffins}),  in the context of a baking recipe, important dimensions are related to the physical properties of butter, margarine and olive oil. When focusing on these type of dimensions butter and margarine seem more similar because they are both solid while olive oil is liquid. In contrast, in the following example (\ref{ex:vegan1}) we bring up ideas about veganism and the dimension of animal versus based plant products becomes very salient. This could bring margarine and olive oil closer together and distance both of them from butter, which is an animal product.

There are important differences between this \emph{salience manipulation} effect and the similarly ``graded'' \emph{contextual modulation} effect. In the previous example (\ref{ex:butter}) \emph{poured} modulated the meaning of the word butter by promoting its ``liquid'' trait. This effect is limited to the word butter. On the contrary, if the context triggers changes in the salience of conceptual dimensions, any word the annotator evaluates after the change takes place will be affected by it. Once the idea of animal vs plant based is introduced, the change takes place in the mind of the annotator and the perception of the meaning of not only butter, but margarine and olive oil is impacted as well. Our hypothesis is that, by using \emph{salience manipulation}, a context like example (\ref{ex:vegan2}) can have a impact in the scoring of the similarity of butter, margarine and olive oil without these words even being present in the context. Something that would be impossible if we were looking only at the \emph{contextual modulation} and \emph{sense selection} effects.

The expectation that priming is the main mechanism for modifying salience has its own implications: \newcite{branigan2000syntactic} found that priming effects are much stronger in the context of as natural dialog as possible, when speakers had no time constraints and could respond at their own pace. These results were taken into account when designing our dataset and annotation methodology: it is crucial for us to create an annotation process in which the annotator interacts with the context, and does so in as natural a way as possible, before they rate the similarity. Because priming is an automatic process, them knowing that they should be annotating similarity in context becomes a lot less important. 

\subsection{Existing Datasets}

There are a few examples of datasets which take context into account. However, so far these have  been motivated by discrete \emph{sense disambiguation}, and therefore take a view of word meaning as discrete (taking one of a finite set of senses) rather than continuous; they are therefore not suited for the more graded effects we are interested to look into. 

The \textbf{Stanford Contextual Word Similarity (SCWS)} dataset \cite{huang2012improving} does contain graded similarity judgements of pairs of words in the context of organically occurring sentences (from Wikipedia). However it was designed to evaluate a discrete multi-prototype model, so the focus was on the contexts selecting for one of the word senses. This resulted in them presenting each of the two words of the pair in their own distinct context. From our point of view this approach has some drawbacks: First, even in the cases where they annotated the same pair twice, we find ourselves with four different contexts, each affecting the meaning of each of the instances of the words independently, and it is not possible to produce a systematic comparison of contextual effects on pairwise similarity. Second, beyond the independent lexical semantics of each word being affected by their independent \emph{local context}, the annotator is being presented with two completely independently occurring contexts at the same time. Even if the two contexts did organically occur on their own, this combination of the two did not, and we have seen before how crucial we think keeping the interaction with the context as natural as possible is. There is no easy way to know how this newly assembled \emph{global context} affects the cognitive state of the annotators and their perception of similarity. The same goes for the contextually-aware models trying to predict their results. Joining the contexts before feeding them to the model could create conflicting, difficult to predict effects, but feeding each context independently is fundamentally different to what humans annotators were presented with.

In addition to these limitations of the independent contexts approach, the scores found in SCWS show a worryingly low inter-rater agreement (IRA), measured as the Spearman correlation between different annotators. As pointed out by \cite{pilehvar2019wic}, the mean IRA between each annotator and the average of the rest, which is considered a human-level upper bound for model's performance, is 0.52; while the performance of a simple context-independent model like word2vec \cite{mikolov2013efficient} is 0.65. Examining the scores more in detail, we find that many scores show a very large standard deviation, with annotators rating  the same pair very differently. One possible reason for this may lie in the annotation design: the task itself does not directly enforce engagement with the context, and the words were presented to annotators highlighted in boldface, making it easy to pick them out from the context without reading it;  thus potentially leading to a lack of engagement of the annotators with the context. 

A lot of these limitations were addressed by the more recent \textbf{Words-in-Context (WiC)} dataset \cite{pilehvar2019wic}. With a more direct and straightforward take on word sense disambiguation, each entry of the dataset is made of two lexicographer examples of the same word. The entry is completed with a positive value (T) if the word sense in the two examples/context is the same, or with a negative value (F) if the contexts point to different word senses. One advantage of this design is that it forces engagement with the context; another is that it creates a task in which context-independent models like word2vec ``would perform no better than a random baseline''. Human annotators are shown to produce healthy inter-rater agreement scores for this dataset. However the dataset is again focused in looking at discrete word senses and cannot therefore capture continuous effects of context in the judgements of similarity between different words.

These datasets are also available only in English, and do not allow models to be evaluated across different languages.
\begin{figure*}[h!]
\begin{tabular}{|llr|} \hline 
\bf Word1: population & 
\bf Word2: people & 
\bf SimLex: $\mu$ 7.68 $\sigma$ 0.80 \\ \hline
\bf Context1 & 
 & 
\bf Context1: $\mu$ 6.49 $\sigma$ 1.40 \\ 
\multicolumn{3}{|p{\linewidth}|}{
Disease also kills off a lot of the gazelle \textbf{population}. There are many \textbf{people} and domesticated animals that come onto their land. If they pick up a disease from one of these domesticated species they may not be able to fight it off and die. Also, a big reason for the decline of this gazelle population is habitat destruction. 
}\\ \hline
\bf Context2 & 
 & 
\bf Context2: $\mu$ 7.73 $\sigma$ 1.77 \\
\multicolumn{3}{|p{\linewidth}|}{
But the discontent of the underprivileged, landless and the unemployed sections remained even after the reforms. The crumbling industries give rise to extreme unemployment, in addition to the rapidly growing \textbf{population}. These \textbf{people} mostly belong to the SC/ST or the OBC. In most cases, they join the extremist organizations, mentioned earlier, as an alternative to earn their livelihoods.
}\\ \hline
\end{tabular}
\caption{Example from the English pilot, showing a word pair with two contexts, each with mean and standard deviation of human similarity judgements. The original SimLex values for the same word pair without context are shown for comparison.}\label{fig:example}
\end{figure*}

\section{Dataset and Task Design}

CoSimLex will be based on pairs of words from SimLex-999 \cite{HillEtAl15SimLex}; the reliability and common use of this dataset makes it a good starting point and allows comparison of judgements and model outputs to the context-independent case. For Croatian and Finnish we use existing translations of Simlex-999 \cite{Mrksic.etal2017,venekoski2017finnish,KittaskThesis2019}. In the case of Slovene, we have produced our own new translation \citelanguageresource{slo-simlex}, following the methodology used by \newcite{Mrksic.etal2017} for Croatian.

The English dataset consists of 333 pairs; the Croatian, Finnish and Slovene datasets of  111 pairs each. Each pair is rated within two different contexts, giving a total of 1554 scores of contextual similarity. This poses a difficult task: to find suitable, organically occurring contexts for each pair; this task is more pronounced for languages with less resources, and as a result the selection of pairs is different for each language.

Each line of CoSimLex will be made of a pair of words selected from Simlex-999; two different contexts extracted from Wikipedia in which these two words appear; two scores of similarity, each one related to one of the contexts; and two scores of standard deviation. Please see Figure~\ref{fig:example} for an example from our English pilot.

\paragraph{Evaluation Tasks and Metrics}

The first practical use of CoSimLex will be as a gold standard for the public SemEval 2020 task 3: \textit{Graded Word Similarity in Context}. The goal of this task is to evaluate how well modern context-dependent embeddings can predict the effect of context in human perception of similarity. In order to do so we define two subtasks and two metrics:

\paragraph{Subtask 1 - Predicting Changes:} 
In subtask 1, participants must predict the \emph{change} in similarity ratings between the two contexts. In order to evaluate it we calculate the difference between the scores produced by the model when the pair is rated within each one of the two contexts. We do the same with the average of the scores produced by the human annotators. Finally we calculate the uncentered Pearson correlation. A key property of this method is that any context-independent model will predict no change and get strongly penalised in this task.  

\paragraph{Subtask 2 - Predicting Ratings:} In subtask 2, participants must predict the absolute similarity rating for each pair in each context. This will be evaluated using Spearman correlation with gold-standard judgements, following the standard evaluation methodology for similarity datasets \cite{HillEtAl15SimLex,huang2012improving}. Good context-independent models could theoretically give competitive results in this task, however we still expect context-dependent models to have a considerable advantage. 

\begin{figure*}[ht]
\begin{tabular}{|llr|} \hline 
\bf Word1: čovjek (adult male) & 
\bf Word2: dijete (child) &
\bf \\  \hline
\bf Context1 & 
 & 
\bf Context1: $\mu$ 2.5 $\sigma$ 1.76 \\ 
\multicolumn{3}{|p{0.975\linewidth}|}{
Špinat ima dosta željeza, ali i oksalne kiseline. Oksalna kiselina veže kalcij i čini ga neupotrebljivim za ljudski organizam. Prema novijim istraživanjima, špinat se ne preporuča kao česta hrana mlađim osobama i \textbf{djeci}, ali je izvrsna hrana za starije \textbf{ljude}.}\\
\multicolumn{3}{|p{0.975\linewidth}|}{
(Spinach has plenty of iron but also oxalic acid. Oxalic acid binds calcium and renders it unusable for the human body. According to recent research, spinach is not recommended as a common food for younger people and \textbf{children}, but it is an excellent food for older \textbf{people}.)}\\ \hline
\bf Context2 & 
 & 
\bf Context2: $\mu$ 4.25 $\sigma$ 0.95 \\
\multicolumn{3}{|p{0.975\linewidth}|}{
Nakon što su \textbf{ljudi} u selu saznali da je trudna, počinju sumnjati na dr. Richardsona jer je on proveo najviše vremena s njom. Kako vrijeme prolazi, pritisak glasina na kraju prisiljava liječnika da se preseli. Odluči se oženiti s Belindom i uzeti \textbf{dijete} sa sobom.}\\
\multicolumn{3}{|p{0.975\linewidth}|}{
(After \textbf{people} in the village find out she is pregnant, they begin to suspect Dr. Richardson because he spent the most time with her. As time goes on, the pressure of the rumors eventually forces the doctor to move. He decides to marry Belinda and take her \textbf{child} with him.)}\\ \hline
\end{tabular}
\caption{Example from the Croatian pilot, showing the word pair with two contexts, each with mean and standard deviation of human similarity judgements. This example showed one of the most significant contextual effects in the pilot; it went in the opposite direction to the one predicted by the expert annotator. Note the effect of stemming: the target word \textit{čovjek} appears in both cases via its irregular plural, \textit{ljudi} (nominative) or \textit{ljude} (accusative); and \textit{dijete} appears in Context 1 in its dative plural form \textit{djeci}. English translations (generated using Google Translate with manual post-correction) are shown here for exposition purposes but are not part of the dataset.}\label{fig:croatian}
\end{figure*}

\section{Annotation Methodology}
As starting point for our annotation methodology, we adapted the annotation instructions used for SimLex-999. This way we benefit from its tested method of explaining how to focus on \emph{similarity} rather than \emph{relatedness} or \emph{association} \cite{HillEtAl15SimLex}. As explained in their original paper, \emph{cup} and \emph{mug} are very similar, while \emph{coffee} and \emph{cup} are strongly related but not similar at all.  For English we adopted a modified version of their crowd-sourcing process:  we use \emph{Amazon Mechanical Turk}, with the same scoring scale (0 to 6), the same post-processing and cleaning of the data (a necessary step when working with this kind of crowd-sourcing platform), and achieve similarly good inter-annotator agreement. For the less-resourced languages, crowdsourcing is not a viable option due to lack of available speakers, and we recruit annotators directly. This means fewer annotators (for Croatian, Finnish and Slovene, 12 annotators vs 27 in English), however the average quality of annotation is a lot higher and the data requires less post-processing - see Section~\ref{sec:status} for details.

\subsection{Finding Suitable Contexts }
For each word pair we need to find two suitable contexts. These contexts are extracted from each language's Wikipedia. They are made of three consecutive sentences and they need to contain the pair of words, appearing only once each. English is by far the easiest language to work with, not only because of the amount and quality of the text contained in the English version of Wikipedia but because the other four languages are highly inflected (Croatian, Finnish and Slovene). To overcome this, we work with data from \cite{11234/1-1989}\footnote{http://hdl.handle.net/11234/1-1989} which contains tokenised and lemmatised versions of Wikipedia for 45 languages.

We first find all the possible candidate contexts for each word pair, and then select those candidates that are most likely to produce different ratings of similarity. The differences are expected to be small, especially in words that don't present several senses and are not highly polysemous, so we need a process that has the most chances of finding contexts that make a difference. We use a dual process in which we use ELMo and BERT to rate the similarity between the target pair within each of the candidate contexts. Then we select the 2 contexts in which ELMo scored the pair as the most similar, and the 2 contexts in which it scored them as most different. We do the same using BERT scores. This gives us 4 contexts in which our target words are scored as very similar by the models and 4 contexts in which they are scored as very different. 

The final selection of two contexts is made by expert human annotators, one per language. We construct online surveys with these 8 contexts and ask them to select the two in which they think the word pair is the most and the least similar, trying to maximise the potential contrast in similarity. In addition, we ask them how much potential for a difference they see in the contexts selected. This gives us not only the contexts we need, but a predicted performance and direction of change for use in later analysis.

In the case of less resourced languages, the smaller size and lower quality of the Wikipedia text resources require some extra steps to ensure the quality of the final annotation. For these languages we run the contexts through a set of heuristic filters to try to remove badly constructed ones. In addition we produce 16 candidates instead of 8 for the expert annotators to choose from, and we add the possibility for them to delete parts of the context in order to make them easier to read. Adding text is not allowed, in order to ensure that contexts are natural.

\subsection{Contextual Similarity Annotation}

The next step is to obtain the contextualised similarity annotations. Our goal is to capture the kind of contextual phenomena discussed in Section~\ref{sec:bg}: lexical meaning modulation and conceptual salience manipulation. In order to maximise our chances we define three goals:
\begin{itemize}
  \item We want the interaction with the context to be as natural as possible,  so as  to maximise priming effects and capture the potential change in the salience of conceptual dimensions.  
  \item We need a way in which annotators have the chance to account for lexical modulation within the sentence. 
  \item We need to avoid the apparent lack of engagement we saw in the SCWS annotators.
\end{itemize}

With these goals in mind we designed a two-step mixed annotation process. Our online survey interface is composed of two pages per pair of words and context (each annotator scores only one of the contexts). In the first page the annotators are presented with the context, and asked to read it and come up with two words ``inspired by it''. Once this is complete, the second page shown presents the context again, but with the pair of words now highlighted in bold; they are now asked to rate the similarity of the pair of words within the sentence.

The second page is the main scoring task; it is designed to capture changes in scores of similarity due both to lexical modulation and --- because we hope the annotators are still primed by their recent previous engagement with the context --- the changes in the salience of conceptual dimensions. The separate task on the first page is intended to make annotators engage fully with the whole context, while maintaining a natural interaction with it to maximise any priming effects.
One of the possible problems we identified in the the SCWS annotation process is the fact that the words were always highlighted in bold, making it easy for annotators (Amazon Mechanical Turk workers) to just look at the pair of words in isolation and to not read the rest of the contexts. Our initial task is designed to prevent this (the words are not bold in the first page).

In English, given the resources available, we follow SimLex-999 closely: we will use Amazon Mechanical Turk to get 27 annotators per pair and context. Annotators do not score the same pair twice: 27 annotators score the pair within one context and another 27 in the other. This means the whole dataset can be annotated at the same time. Reliability of annotations will be ensured by an adapted version of SimLex-999's post-processing, which includes rating calibration and the filtering of annotators with very low correlation to the average rating. In addition, we will use responses to the first annotation question to check annotator engagement with the context text and thus filter low quality raters.

For Croatian, Finnish and Slovene  we recruit annotators directly: this means we have less of them (12 vs 27) but we expect the quality of the annotation to be better (and pilots confim this -- see below). It also means, howeve, that we must use the same annotators to rate the two contexts of each pair. This has an avantage, because it controls for the variation in the particular judgement of different annotators, but means that we introduce a week's delay in between annotations in order to make sure they don't remember, and are influenced by, their own previous score.

\begin{table*}[p]
\begin{center}
\begin{tabular}{|l|l|r|r|r|r|l|}
\hline

\textbf{Word1} & \textbf{Word2} & \textbf{Context1} & \textbf{Context2} & \textbf{STDev1} & \textbf{STDev2} & \textbf{P-Value} \\ \hline
water & ice & 2.57 & 8.13 & 2.60 & 1.82 & 2.18E-08$^{\ast\ast}$\\ \hline
friendly & generous & 4.44 & 3.92 & 2.85 & 3.56 & 0.258 \\ \hline
keep & protect & 2.50 & 3.75 & 2.66 & 2.22 & 0.036$^{\ast\ast}$\\ \hline
pact & agreement & 8.73 & 8.97 & 1.89 & 1.53 & 0.302 \\ \hline
narrow & broad & 0.42 & 1.97 & 1.19 & 2.60 & 0.012$^{\ast\ast}$ \\ \hline
arm & neck & 3.81 & 1.27 & 2.89 & 1.97 & 0.002$^{\ast\ast}$ \\ \hline
cottage & cabin & 8.07 & 9.56 & 2.37 & 0.94 & 0.003$^{\ast\ast}$ \\ \hline
inform & notify & 9.31 & 9.80 & 0.97 & 0.55 & 0.019$^{\ast\ast}$ \\ \hline
mother & guardian & 3.94 & 7.28 & 3.15 & 2.54 & 0.0001$^{\ast\ast}$ \\ \hline
car & bicycle & 4.12 & 4.85 & 2.58 & 2.46 & 0.169 \\ \hline

\end{tabular}
\caption{Example results from the English dataset showing the mean similarity, standard deviation and p-value calculated using the Mann-Whitney U test.}
\label{tab:eng-example}
\end{center}

\begin{center}
\begin{tabular}{|l|l|r|r|r|r|l|}
\hline

\textbf{Word1} & \textbf{Word2} & \textbf{Context1} & \textbf{Context2} & \textbf{STDev1} & \textbf{STDev2} & \textbf{P-Value} \\ \hline
nadbiskup & biskup & 6.67 & 6.30 & 2.66 & 2.61 & 0.325 \\ \hline
sretan & mlad & 1.50 & 0.30 & 2.28 & 0.67 & 0.099$^{\ast}$ \\ \hline
kost & čeljust & 6.00 & 3.33 & 2.38 & 2.24 & 0.013$^{\ast\ast}$ \\ \hline
zvijer & životinja & 9.44 & 6.30 & 1.09 & 3.09 & 0.004$^{\ast\ast}$ \\ \hline
priča & tema & 2.59 & 7.64 & 1.88 & 2.51 & 0.0004$^{\ast\ast}$ \\ \hline

\end{tabular}
\caption{Example results from the Croatian dataset showing the mean similarity, standard deviation and p-value calculated using the Mann-Whitney U test.}
\label{tab:hr-example}
\end{center}
\end{table*}

\begin{figure*}[p]
\includegraphics[width=\textwidth]{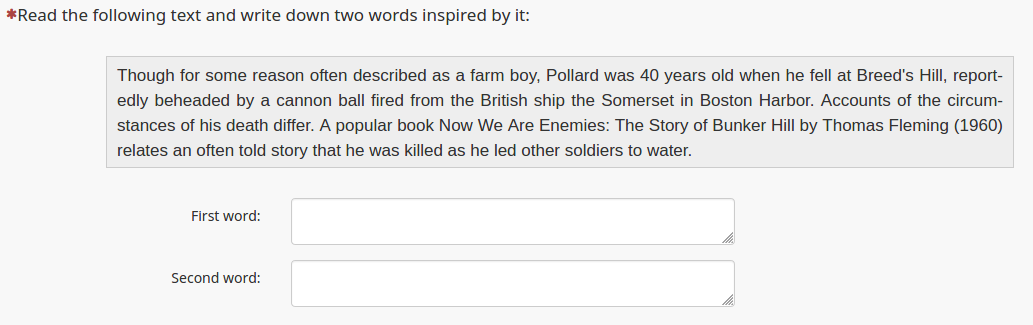}
\caption{First page shown for each word pair annotation task: annotators must read the context and come up with two words inspired by it. At this point, the word pair to be scored is not known to the annotator.}
\label{fig:page1}
\vspace*{0.5cm}
\includegraphics[width=\textwidth]{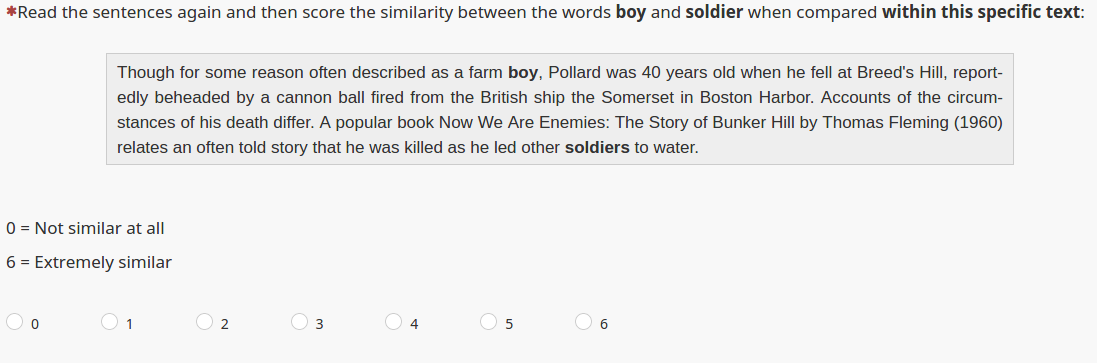}
\caption{Second page shown for each word pair annotation task: the same context is now shown with the target words in bold, and annotators must give a similarity score for the word pair within that particular context.}
\label{fig:page2}
\end{figure*}

\clearpage

\section{Current Status}\label{sec:status}

\paragraph{Methodology prototyping}
We have run three pilots with 13 pairs of words each to confirm the annotation design and methodology. 
Each study tested a slight variation: in the first pilot, annotators rated \emph{relatedness} in addition to similarity; the second focused on similarity, and tested the use of contexts related to the target words but not containing them; the third experimented with marking the target words in the context paragraphs using boldface font.

The first pilot confirmed that (as with SimLex) similarity is a more useful metric for this task than relatedness, displaying a higher inter-annotator agreement and more variation between contexts; we therefore use similarity as the basis of our dataset, as described above.

The results of the second pilot saw significant contextual effects in many examples, including some in which the target words weren't included in the contexts. This indicates that our method seems suitable for capturing priming effects and salience manipulation, or at least some kind of cognitive effects different from lexical contextualisation. 

The third pilot showed much lower agreement and lower difference between contexts: we take this as confirmation of our suspicion (from analysis of SCWS) that marking the target words makes it easy for annotators to ignore the rest of the context paragraph, and therefore use the two-stage annotation methodology described above, in which target words are  \emph{not} initially marked.

\paragraph{Results}

The dataset contains 341 entries in English, 113 in Croatian, 112 in Slovene and 25 in Finnish. Each of the entries contains a pair of words evaluated in two different contexts. Please see Table~\ref{tab:eng-example} for examples from the English results and Table~\ref{tab:hr-example} for Croatian.

Inter-rater agreement (IRA) is measured as Spearman correlation between each rating and the average values. After post-processing the data our dataset's IRA is surprisingly similar between the different languages: for English, Croatian and Slovene the mean is $\rho$ = 0.77, while Finnish achieves $\rho$ = 0.80 (likely due to the small sample); these compare well to other related datasets (SimLex-999 $\rho$ = 0.78, SCWS $\rho$ = 0.52). The crowd-sourced nature of English data results in a higher percentage of annotations being dropped in the post-processing. However the fact that both methods converge to the same IRA is encouraging, and seems to indicate that both methods achieve comparable results.     

The statistical significance of the difference in similarity evaluation between contexts was assessed using the Mann-Whitney U test. In English, from the 341 entries, 220 results showed a statistically significant difference at $p<0.1$ (ratio = 0.65), and 208 did so at $p<0.05$ (ratio = 0.61). The results again where quite similar for Croatian and Slovene with 73 and 65 statistically significant results at $p<0.1$ (ratio = 0.65 and 0.58). Finnish results showed a much smaller ratio of statistically significant results (8 entries, ratio = 0.33), which could be due to the small sample but may deserve further investigation. However, as in the case of the inter-rater agreement, the fact that English, Croatian and Slovene results are very similar is a good sign for both methodologies: the English crowd-sourced annotation and the smaller sample of better quality annotation we used for Croatian and Slovene.


\section{Conclusion}

The growing use of context-dependent language models and representations in NLP motivates the need for a dataset against which they can be evaluated, and which can test their ability to reflect human perceptions of context-dependent meaning. CoSimLex will provide such a dataset, and do so across a number of less-resourced languages as well as English.
The full dataset was provided for the evaluation stage of SemEval 2020 at the beginning of February 2020, and will be made public when the competition is over (before the LREC2020 conference).

\section{Acknowledgements}
This research is supported by the European Union’s Horizon 2020 research and innovation programme under grant agreement No 825153, project EMBEDDIA (Cross-Lingual Embeddings for Less-Represented Languages in European News Media). The results of this publication reflect only the authors' views and the Commission is not responsible for any use that may be made of the information it contains.
The first author is also supported by the EPSRC and AHRC Centre for Doctoral Training in Media and Arts Technology (EP/L01632X/1); the second author is also supported by the EPSRC project Streamlining Social Decision Making for Improved Internet Standards (SoDeStream, EP/S033564/1).


\clearpage

\section{Bibliographical References}
\label{ref:main}
\bibliographystyle{lrec}
\bibliography{CoSimLex}

\section{Language Resource References}
\label{ref:lr}
\bibliographystylelanguageresource{lrec}
\bibliographylanguageresource{languageresource}

\end{document}